\documentclass[runningheads]{llncs}
\pdfoutput=1 
\usepackage{amssymb, amsmath, dblfloatfix, textcomp, cite, multirow, tabularx, color}
\usepackage{url,graphbox, afterpage, floatrow, booktabs}
\DeclareGraphicsExtensions{.pdf,.jpeg,.png,.tif,.tiff}
\usepackage{graphicx}
\graphicspath{{figures/}}

\usepackage[export]{adjustbox}
\usepackage{float}
\usepackage{subfig}

\newfloatcommand{capbtabbox}{table}[][\FBwidth]

\usepackage{marvosym}
\newcommand{\envelope}{(\raisebox{-.5pt}{\scalebox{1.45}{\Letter}}\kern-1.7pt)}
\newsavebox{\bigleftbox}

\begin{document}
\mainmatter  

\title{Automated Performance Assessment in Transoesophageal Echocardiography with Convolutional Neural Networks}

\titlerunning{Automated Performance Assessment in TEE with CNNs}
\author{Evangelos B. Mazomenos\inst{1{\envelope}} \and Kamakshi Bansal\inst{1} \and Bruce Martin\inst{3} \and Andrew Smith\inst{3} \and Susan Wright\inst{2} \and Danail Stoyanov\inst{1{\envelope}}}

\authorrunning{E. B. Mazomenos et al.}

\institute{UCL Wellcome/EPSRC Centre for Interventional and Surgical Sciences,\\
Department of Computer Science, University College London, London, U.K. \\ \email{\{e.mazomenos, danail.stoyanov\}@ucl.ac.uk} \and
St George’s University Hospitals, NHS Foundation Trust, London, U.K.
\and St Bartholomew’s Hospital, NHS Foundation Trust, London, U.K.
}
\maketitle 
\begin{abstract}
Transoesophageal echocardiography (TEE) is a valuable diagnostic and monitoring imaging modality. Proper image acquisition is essential for diagnosis, yet current assessment techniques are solely based on manual expert review. This paper presents a supervised deep learning framework for automatically evaluating and grading the quality of TEE images. To obtain the necessary dataset, 38 participants of varied experience performed TEE exams with a high-fidelity virtual reality (VR) platform. Two Convolutional Neural Network (CNN) architectures, AlexNet and VGG, structured to perform regression, were finetuned and validated on manually graded images from three evaluators. Two different scoring strategies, a criteria-based percentage and an overall general impression, were used. The developed CNN models estimate the average score with a root mean square accuracy ranging between 84\%-93\%, indicating the ability to replicate expert valuation. Proposed strategies for automated TEE assessment can have a significant impact on the training process of new TEE operators, providing direct feedback and facilitating the development of the necessary dexterous skills.
\end{abstract}

\section{Introduction}
\label{sec1}
Transoesophageal echocardiography (TEE) is the standard for anaesthesia management and outcome evaluation in cardiovascular interventions. It is also used extensively for monitoring critically ill patients in intensive care. The success of the procedure is chiefly dependent on the acquisition of appropriate US views that allow for a thorough hemodynamic evaluation to be conducted. To capture high-quality TEE images, practitioners must possess refined psychomotor abilities and advanced hand-eye coordination. Both require rigorous training and practice. 

To facilitate the education of new interventionalists, standardize reporting and quality, accreditation organisations have defined a set of practice guidelines, for performing a comprehensive TEE exam \cite{hahn13, flachskampf11}. Nevertheless, training is hindered because performance evaluation is, almost exclusively, carried out through expert supervision. Typically, senior medical personnel grade TEE exams and review logbooks, a laborious process that requires significant amount of time. As a result, trainees rarely receive immediate feedback. Performance evaluation is a key element in interventional medicine and alternative, preferably automated, methods for evaluating TEE competency are necessary \cite{song12}. So far, objective assessment in TEE is focused exclusively on the kinematic analysis of the US probe with various motion parameters found to be indicative of the level of operational expertise \cite{matyal15, mazomenos16mo}. Although these are important findings, probe kinematic information is not available in clinical settings and only captured in simulation systems. Recent studies emphasise the benefits of virtual reality (VR) simulators that offer a risk-free environment where trainees can practice repeatedly at the their own convenience \cite{bose11}. Evidence of performance improvement after training on VR systems, as well as skill retention and transferability have been reported \cite{ferrero14, damp13, prat16, sohmer14, arntfield15}. Incorporating performance evaluation and structured feedback, will allow further use of VR platforms for training and assessment. 

In this work, we introduce the use of Convolutional Neural Networks (CNNs) for the automated evaluation of acquired TEE images. CNNs have found many applications in medical imaging and computer-assisted surgery \cite{litjens17}, but this is the first time they are applied to skills assessment. We aim to generate high-level features in order to develop a system capable of assigning TEE performance scores, essentially replicating expert evaluation. We generated a dataset of 16060 simulated TEE images from participants of varied experience and use it to retrain two CNN architectures (Alexnet, VGG), converted to perform regression. Three reviewers provided ground truth labels by blindly grading the images with two different manual scores. Tested on a set of 2596 images, the developed CNN architectures estimated the average reviewers' score with a root mean square error (RMSE) ranging from 7\%-14\%. This level of accuracy, which is near the resolution of the average scores from the three evaluators, highlights the potential of CNN algorithms for refined performance evaluation.

\section{Methods}
\label{sec2}
\subsection{Dataset generation}

\begin{figure}[!tb]
  \centering
  \sbox{\bigleftbox}{%
    \begin{minipage}[b]{.6\textwidth}
      \subfloat[] {\includegraphics[width=1.175\textwidth, keepaspectratio]{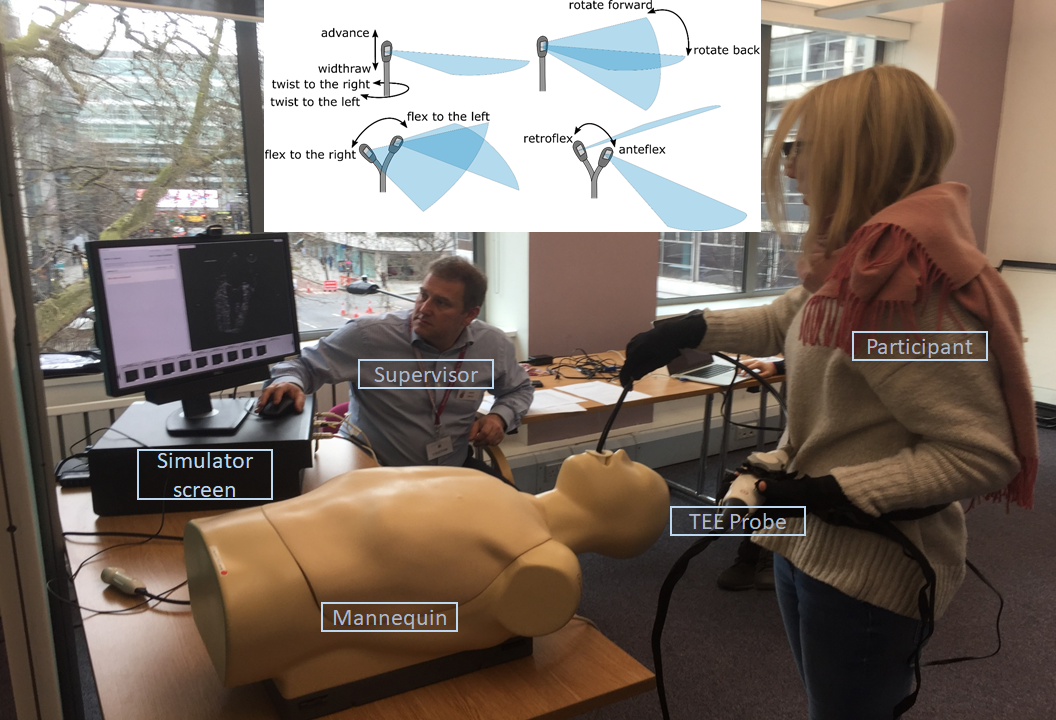} \label{fig1a} }  \end{minipage}%
      }
      \usebox{\bigleftbox}%
      \begin{minipage}[b][\ht\bigleftbox][s]{0.5\textwidth}
      \centering
      \subfloat[] {\includegraphics[width=0.521\textwidth, keepaspectratio]{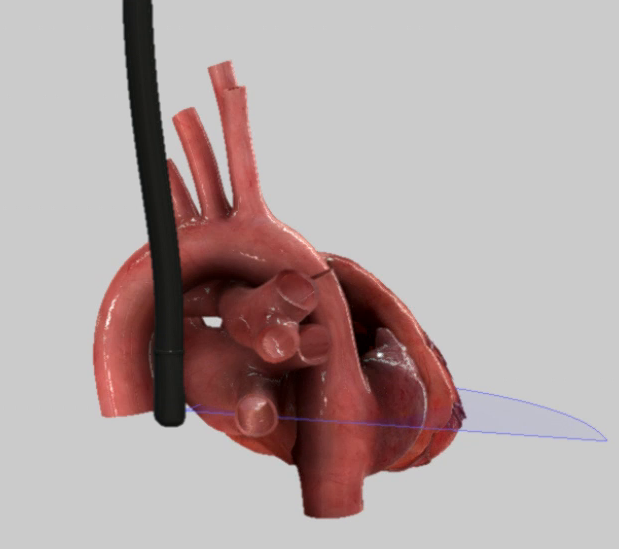} \label{fig1b}}
      \vspace{1pt}
	   \subfloat[] {\includegraphics[width=0.521\textwidth, keepaspectratio]{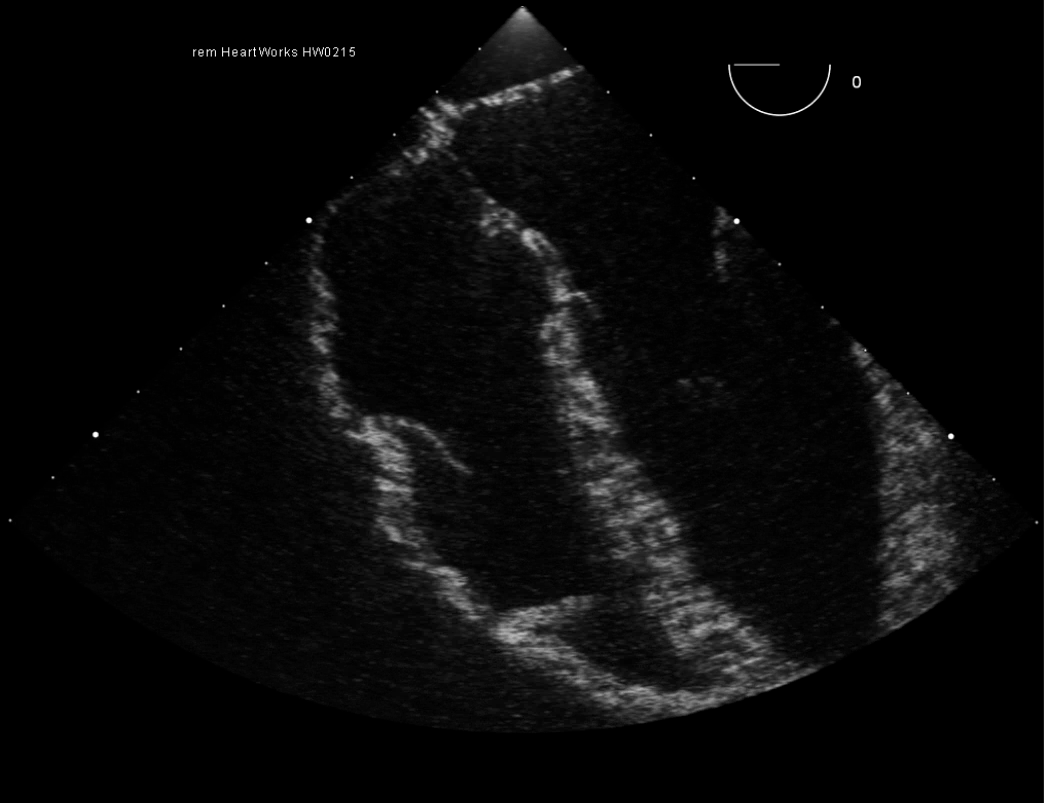} \label{fig1c}}
     \end{minipage}
     \vspace*{-10pt}
     \caption{(a) The HeartWorks simulator, inset the US probe movements; (b) The heart model, the probe and US scanning field; (c) The simulated TEE image}
	\label{fig1}
\end{figure}
\begin{figure}[!tb]
 \begin{tabular}[b]{@{}l@{}l@{}l@{}l@{}l@{}}
\parbox[b]{2.5cm}{\centering 1: ME4C (TV)} & 
\parbox[b]{2.5cm}{\centering 2: ME2C} & 
\parbox[b]{2.5cm}{\centering 3: ME AV SAX} & 
\parbox[b]{2.5cm}{\centering 4: TG mid SAX} & 
\parbox[b]{2.5cm}{\centering 5: ME RV\\\vspace{-2pt} inflow-outflow} \\%
\subfloat{\includegraphics[align=l, height=2.2cm, width=2.2cm]{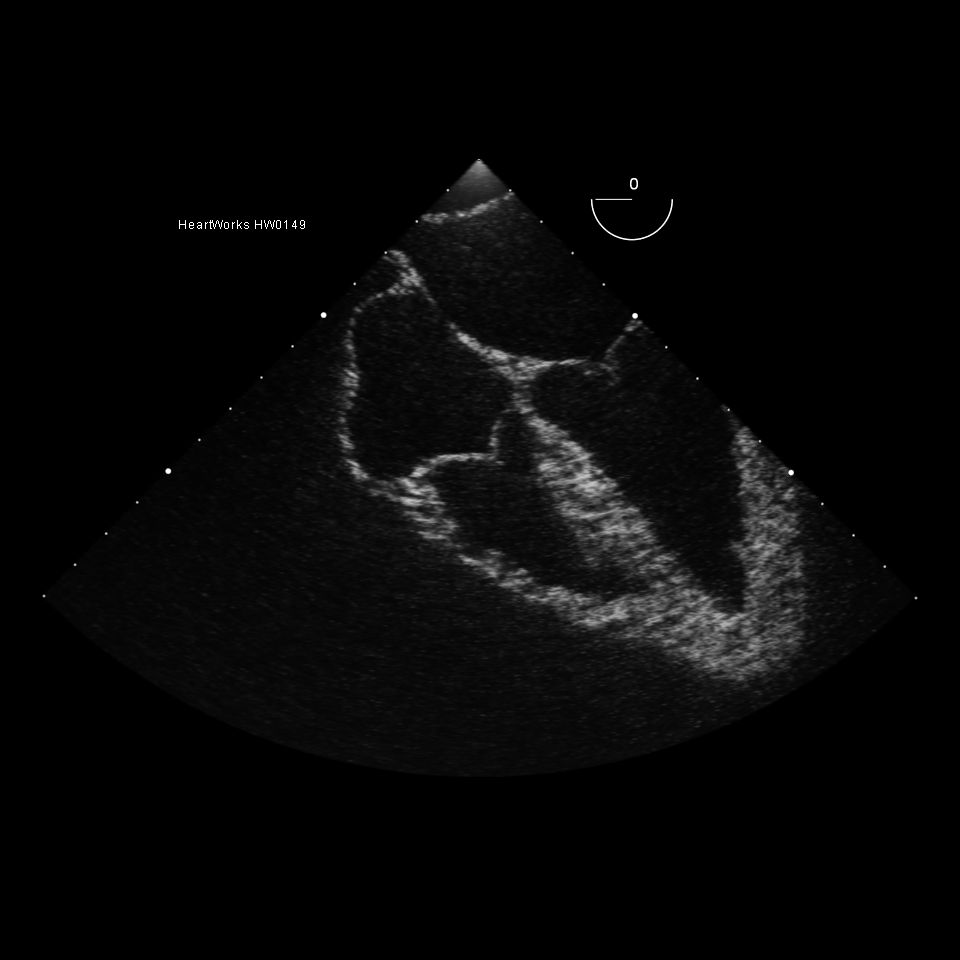}} & 
\subfloat{\includegraphics[align=l, height=2.2cm, width=2.2cm]{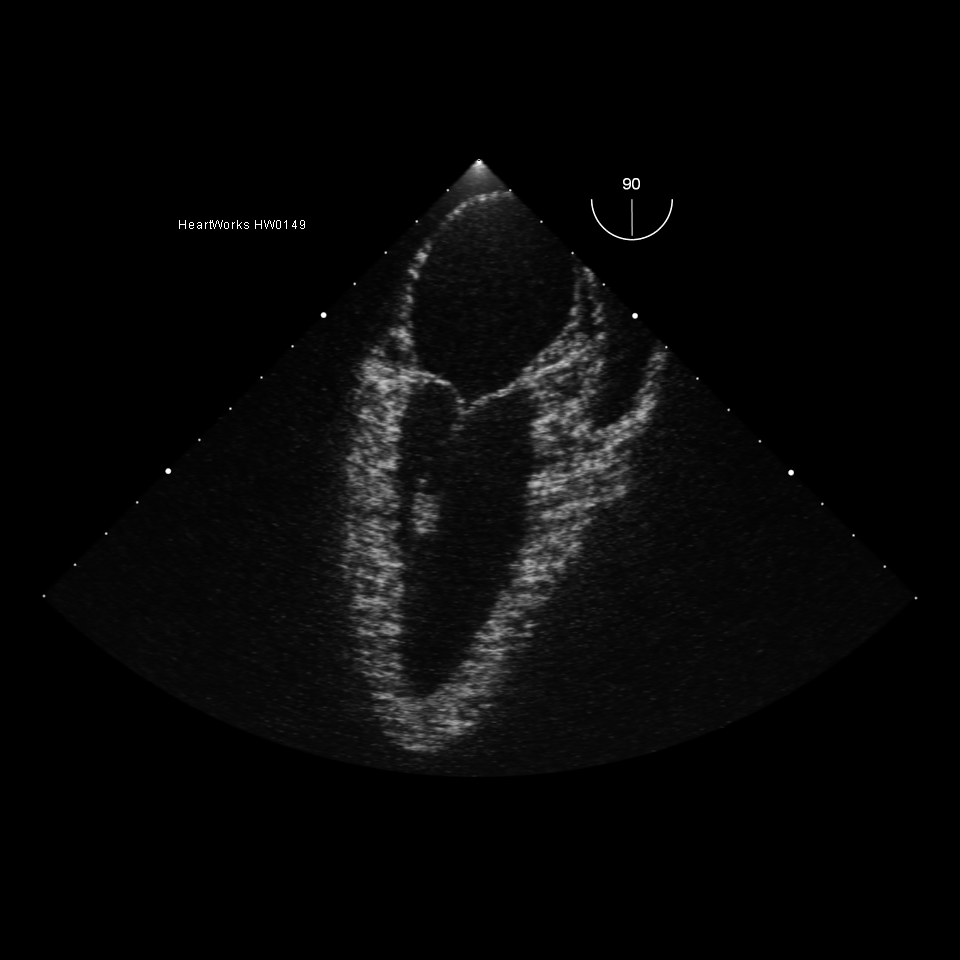}} & 
\subfloat{\includegraphics[align=l, height=2.2cm, width=2.2cm]{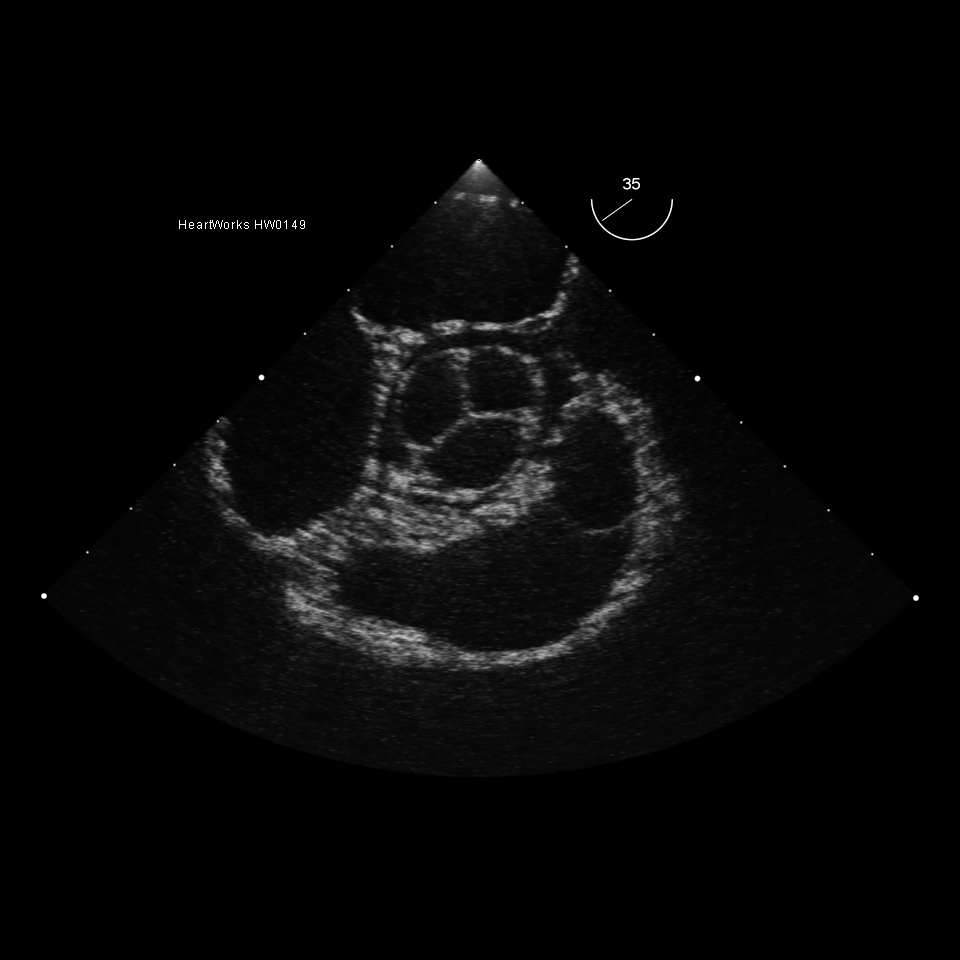}} & 
\subfloat{\includegraphics[align=l, height=2.2cm, width=2.2cm]{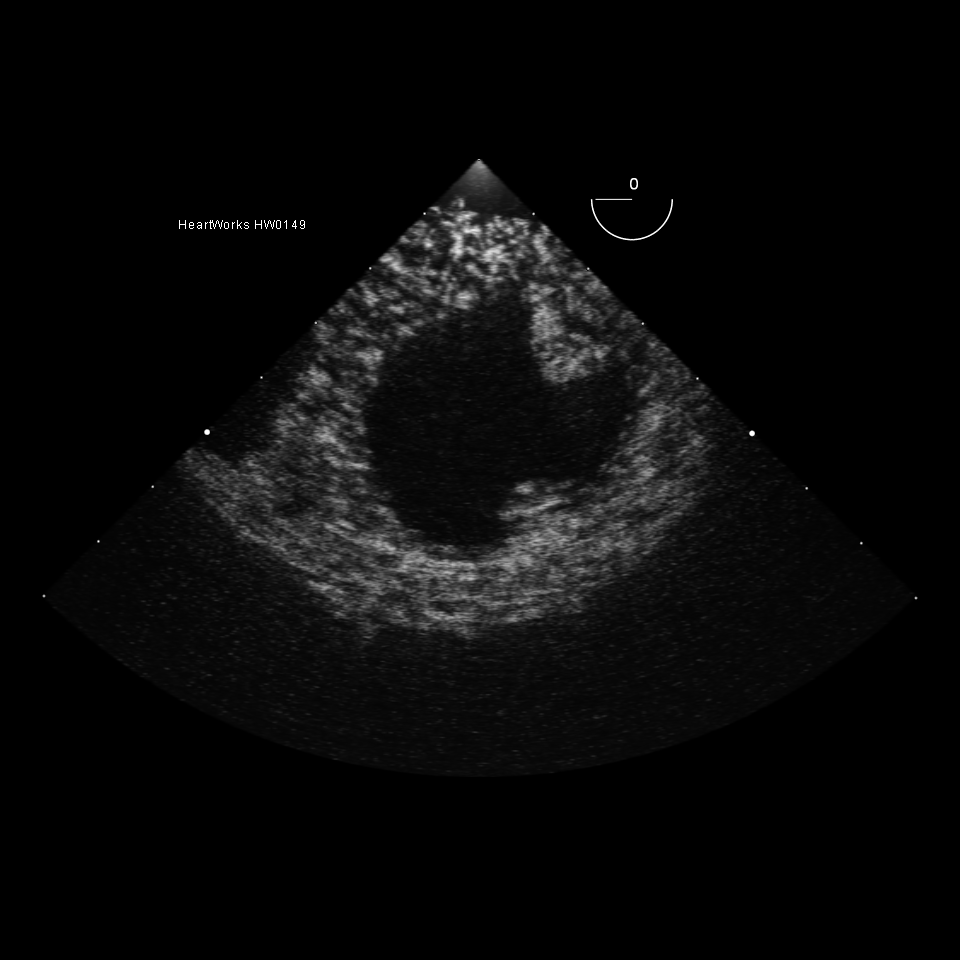}} & 
\subfloat{\includegraphics[align=l, height=2.2cm, width=2.2cm ]{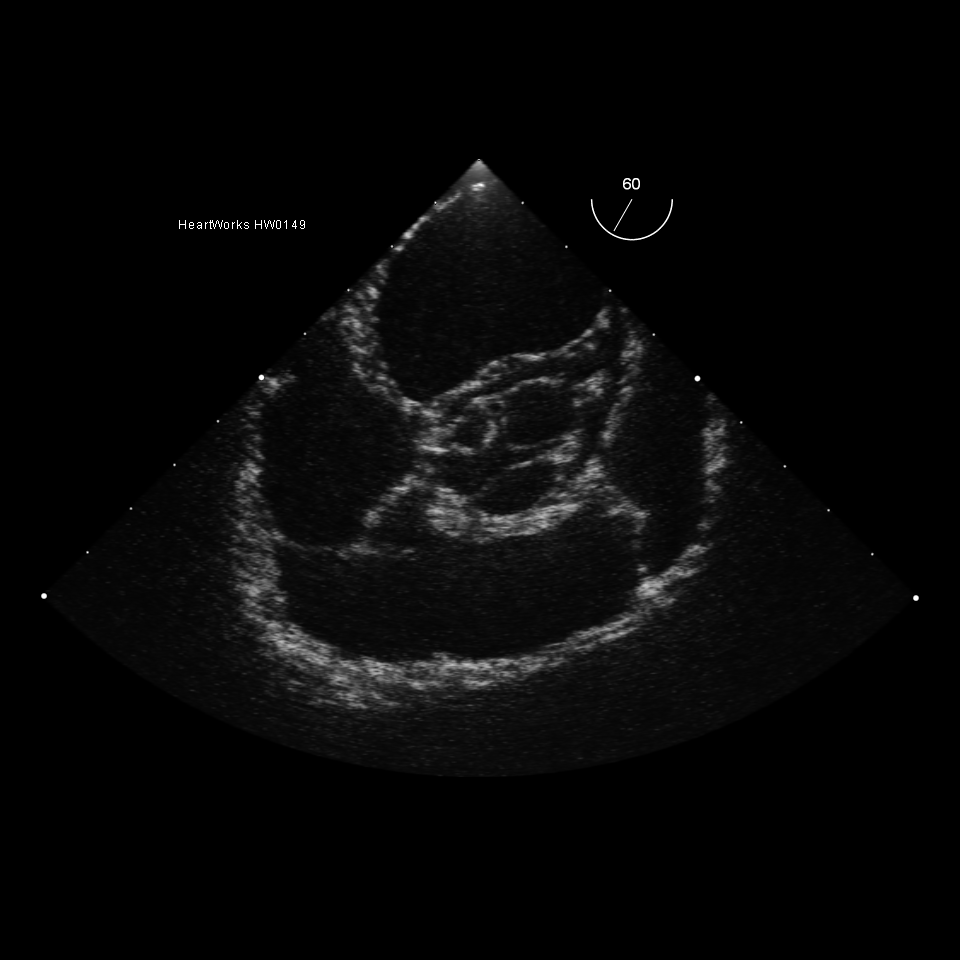}} \\%
\parbox[b]{2.5cm}{\centering  6:ME AV LAX} & 
\parbox[b]{2.5cm}{\centering  7: TG2C} &  
\parbox[b]{2.5cm}{\centering  8: ME4C (LV)} &  
\parbox[b]{2.5cm}{\centering  9:dTG LAX} &  
\parbox[b]{2.5cm}{\centering 10: ME MV\vspace{-2pt}\\commis} \\%
\subfloat{\includegraphics[align=l, height=2.2cm, width=2.2cm]{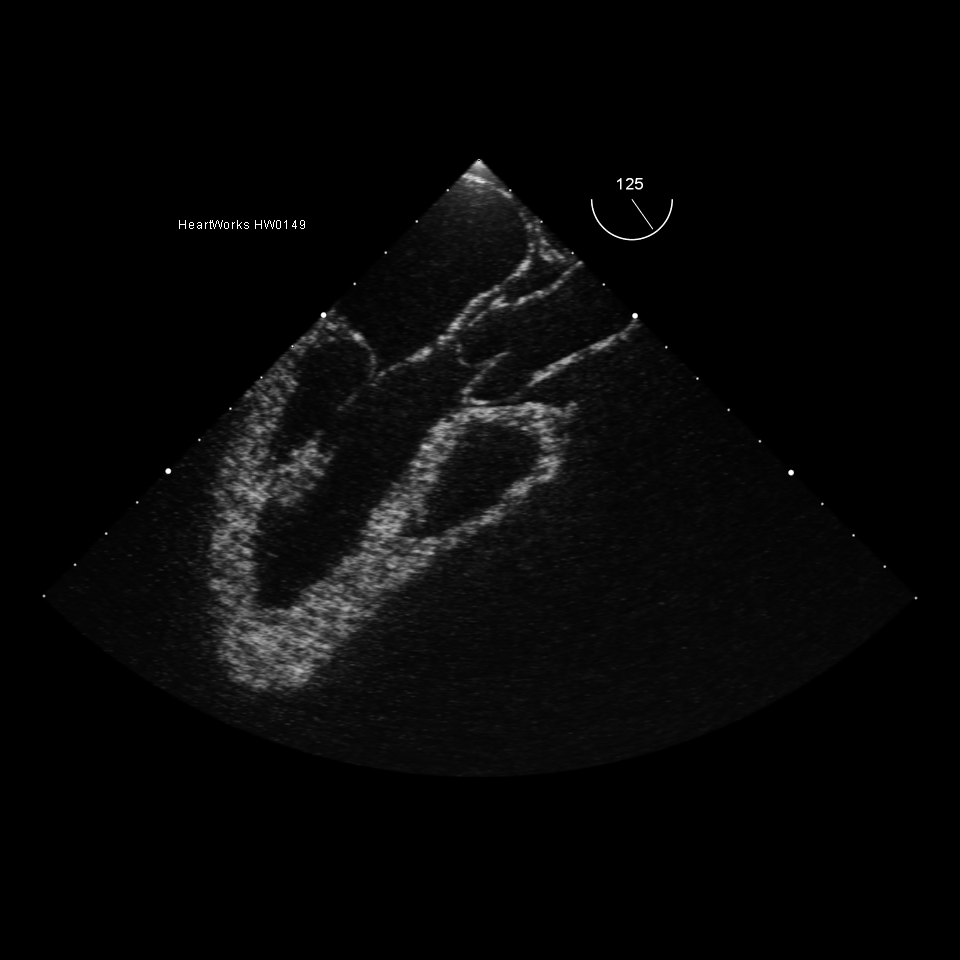}} & 
\subfloat{\includegraphics[align=l, height=2.2cm, width=2.2cm]{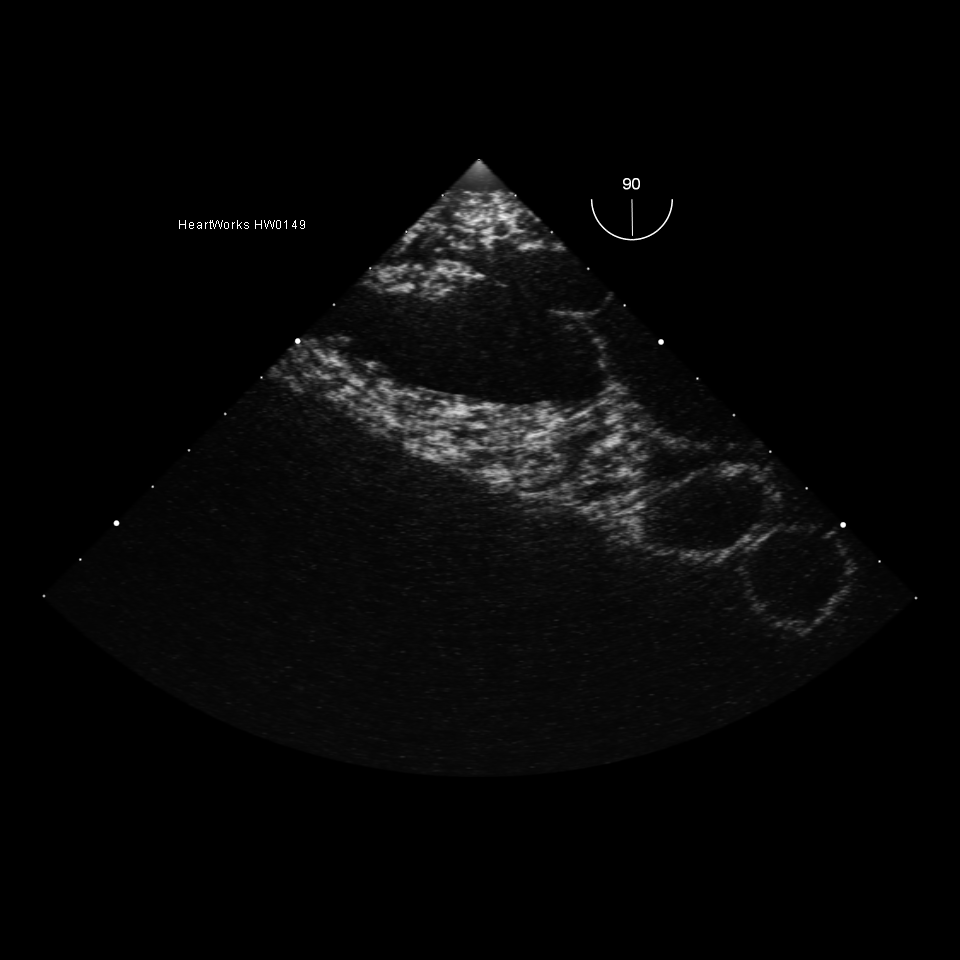}} & 
\subfloat{\includegraphics[align=l, height=2.2cm, width=2.2cm]{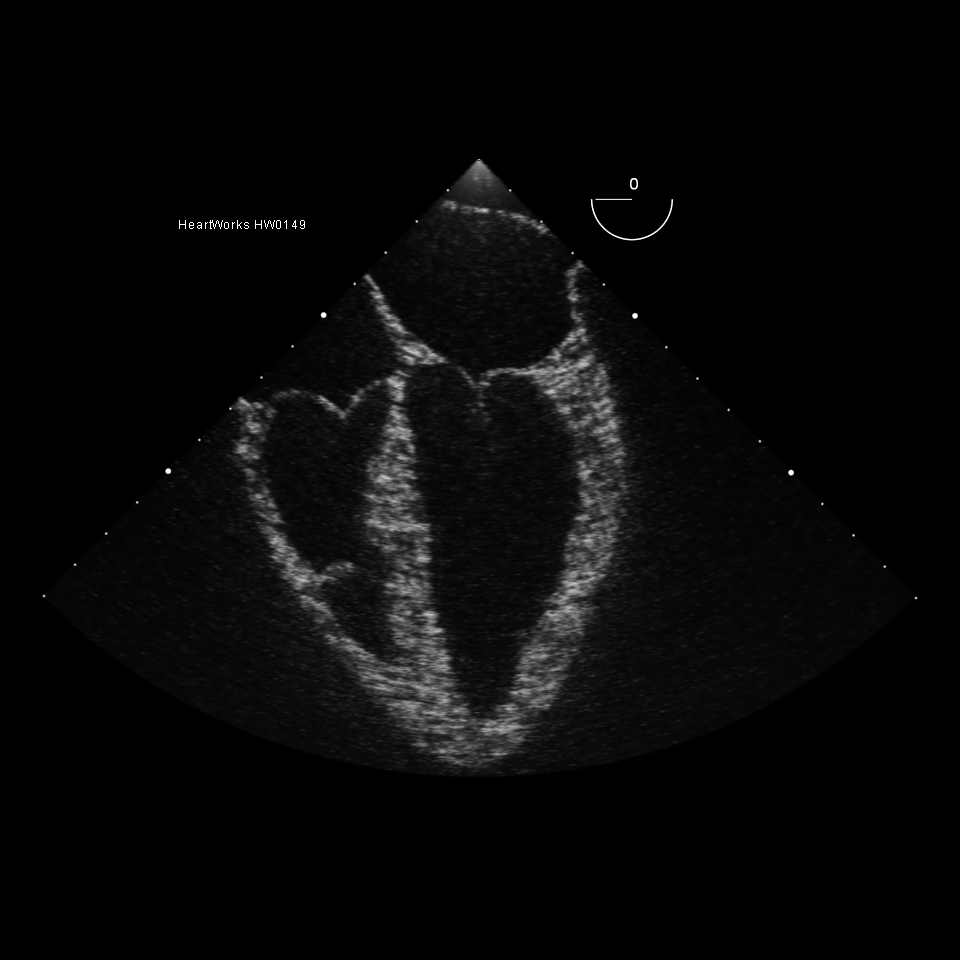}} & 
\subfloat{\includegraphics[align=l, height=2.2cm, width=2.2cm ]{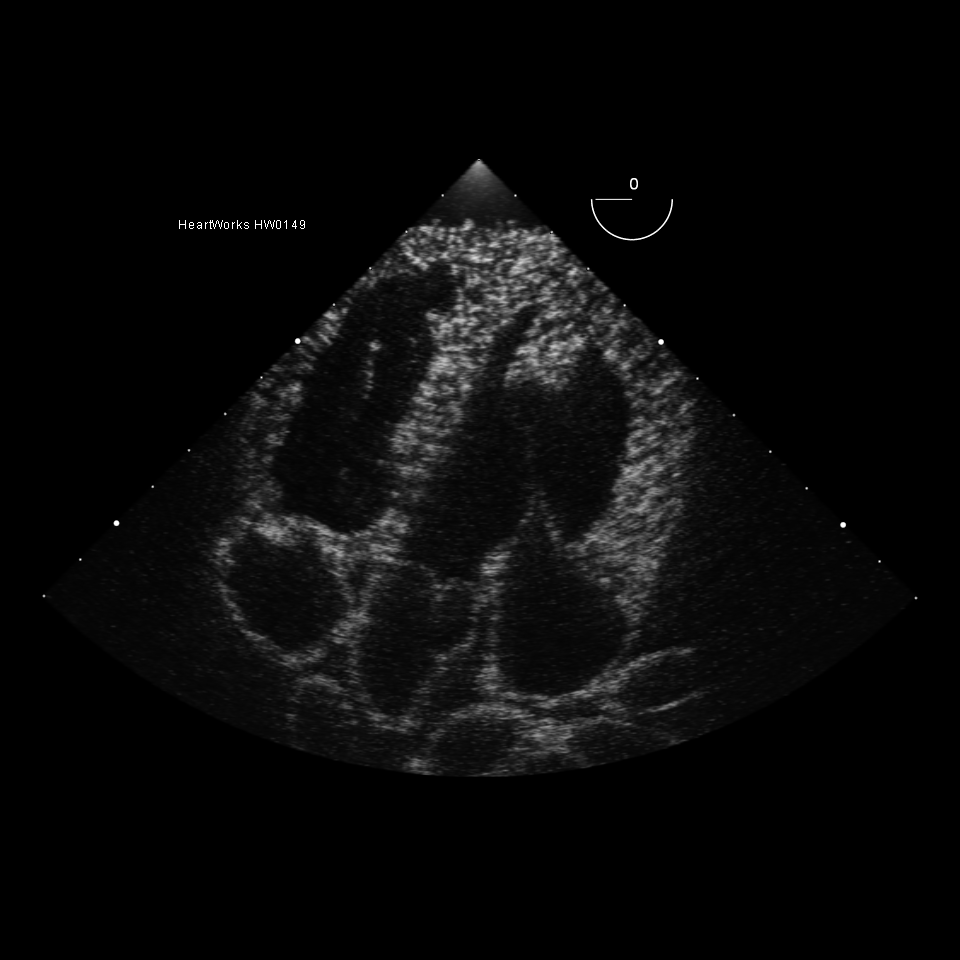}} & 
\subfloat{\includegraphics[align=l, height=2.2cm, width=2.2cm]{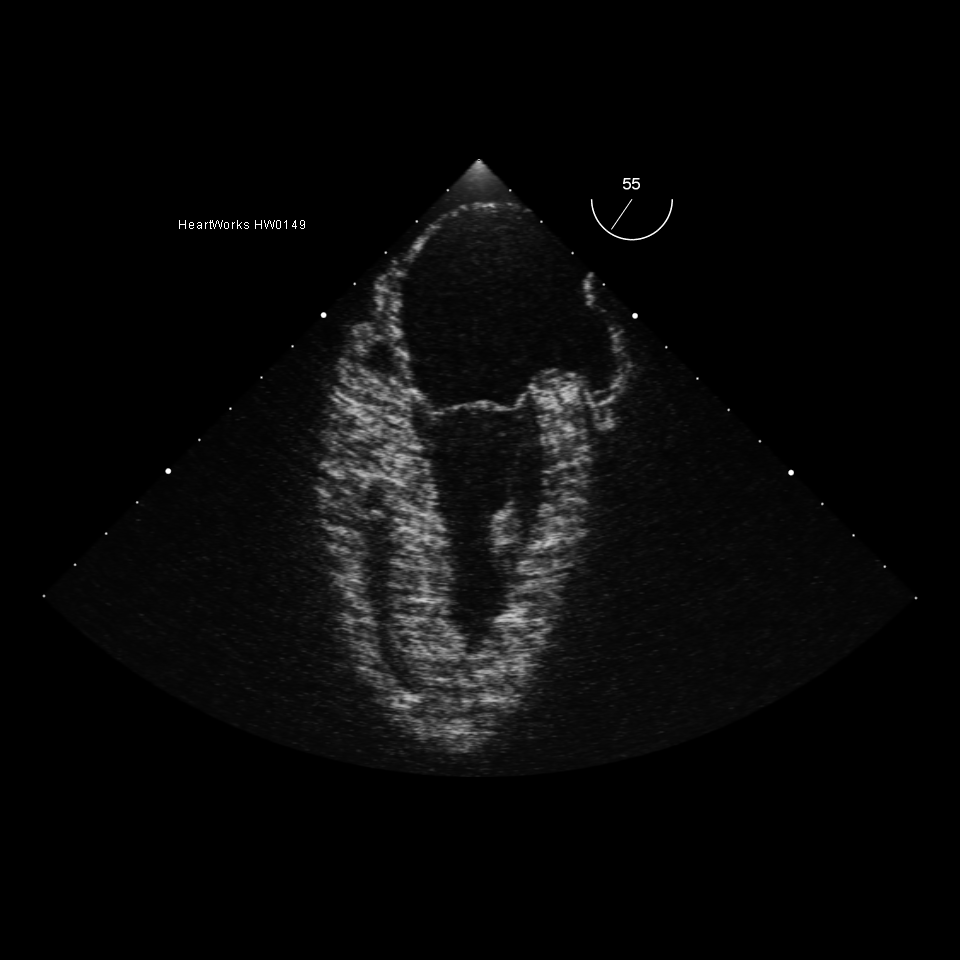}} 
\end{tabular}
\caption{\small The sequence of the 10 TEE views used in the study: 1: Mid-Esophageal 4-Chamber (centred at tricuspid valve), 2: Mid-Esophageal 2-Chamber, 3: Mid-Esophageal Aortic Valve Short-Axis, 4: Transgastric Mid-Short-Axis, 5: Mid-Esophageal Right Ventricle inflow-outflow, 6: Mid-Esophageal Aortic Valve Long-Axis, 7: Transgastric 2-Chamber, 8: Mid-Esophageal 4-Chamber (centred at left ventricle), 9: Deep Transgastric Long-Axis, 10: Mid-Esophageal Mitral Commissural.}
\label{fig2}
\end{figure}

We experimented using the HeartWorks TEE simulation platform, (Inventive Medical, Ltd, London, U.K.) a high-fidelity VR simulator that emulates realistic exam settings (\figurename\,\ref{fig1}). Synthetic US images are generated based on an anatomically accurate cardiac model, illustrated in \figurename\,\ref{fig1b}, that is deformable to mimic a beating heart. A detector on the probe's tip extracts the position and orientation of the US scanning field which are then used to graphically render the 2D US slice (\figurename\,\ref{fig1c}) from the 3D model. The data collection study consisted of a single TEE exam in which participants had to capture 10 US views, shown in \figurename\,\ref{fig2} in a specific sequence. The selected views are a subset of the 20 suggested views recommended by ASE/SCA \cite{hahn13} and include planes from every depth window of the TEE exam (mid-esophageal, transgastric and deep-transgastric)). Experiments were performed under supervision by a consultant anaesthetist that introduced the study and relayed the sequence of views. For capturing and storing data the participant used a foot-pedal to generate a full-HD image and a short video ($\sim 1.5s$) of the imaged US plane. Each video contained 44 frames. 

\begin{figure}[!ht]
  \TopFloatBoxes
  \begin{floatrow}
  \ffigbox[\FBwidth][]{%
  \begin{tabular}{@{}ll@{}}        
  \subfloat{\includegraphics[scale=0.33]{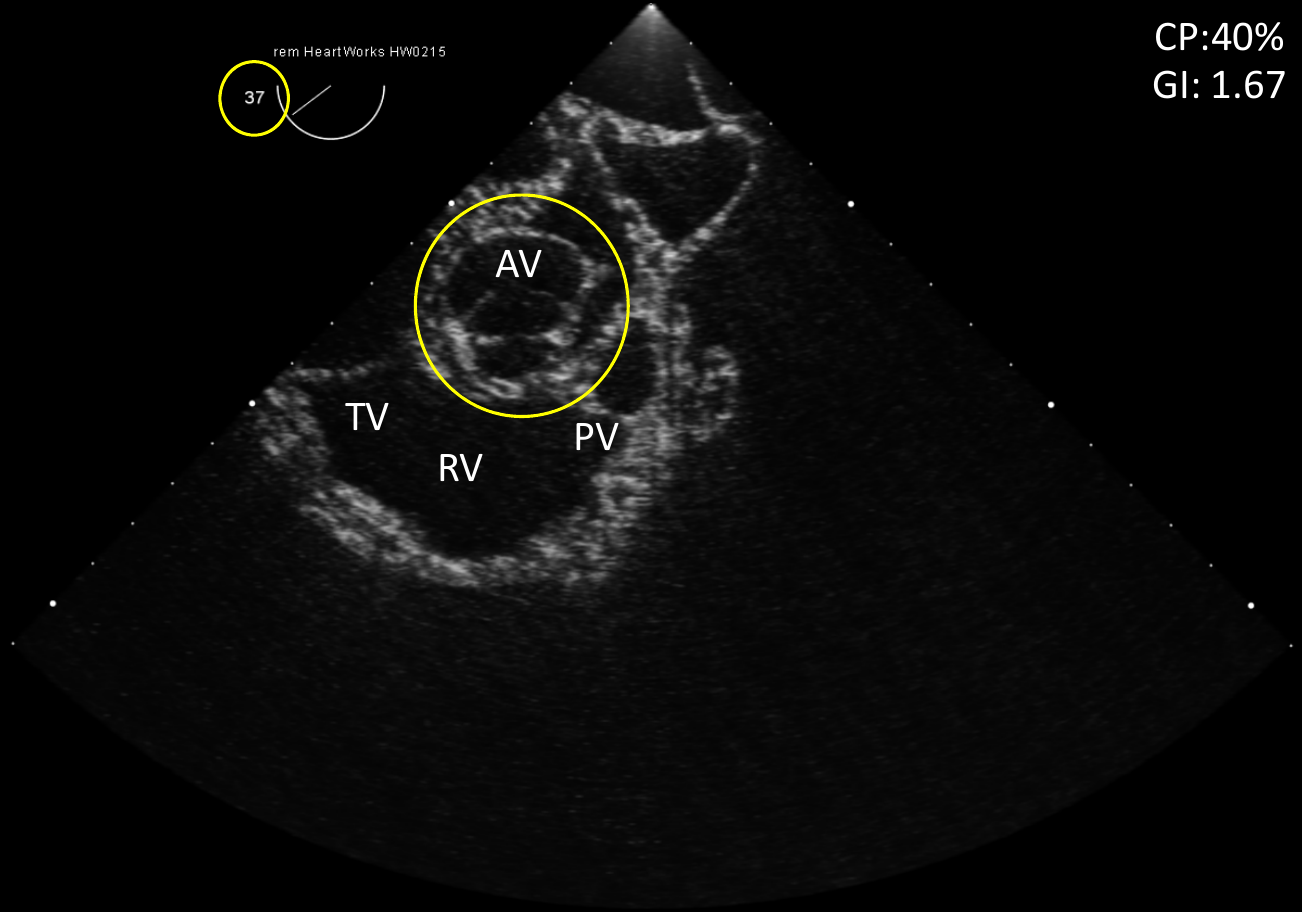}} & \subfloat{\includegraphics[scale=0.33]{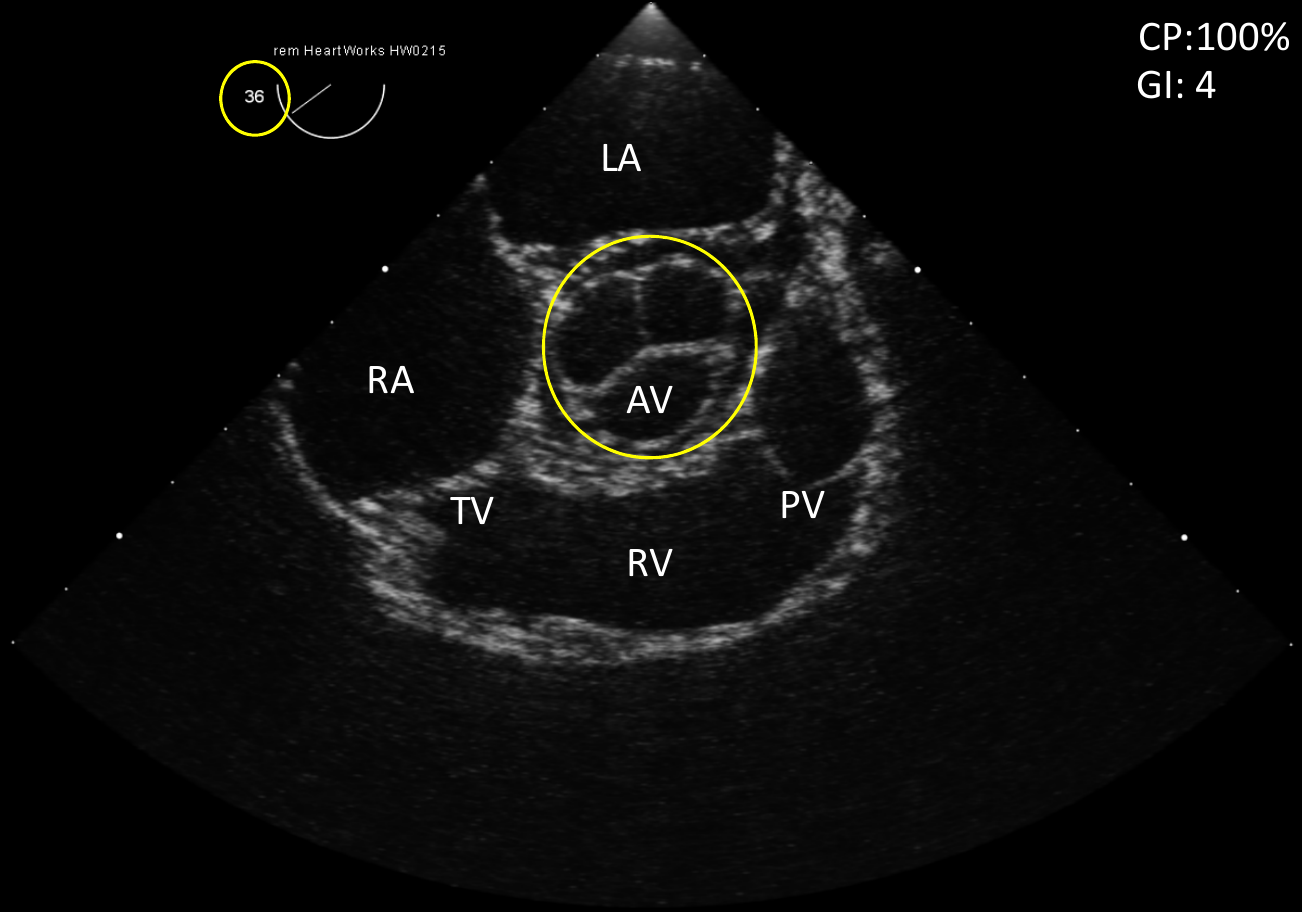}} \\
  \\
  \subfloat{\includegraphics[scale=0.33]{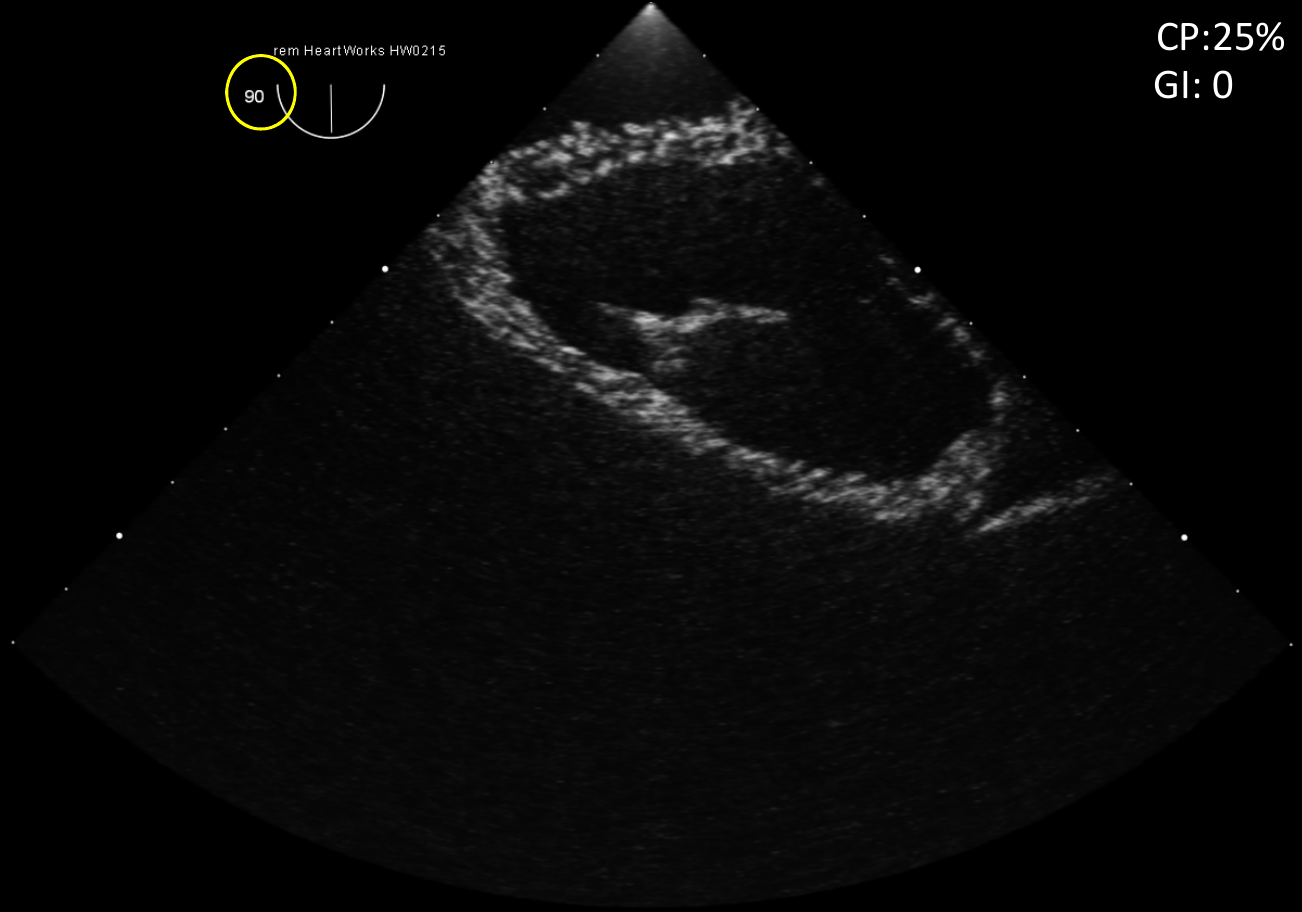}}  & \subfloat{\includegraphics[scale=0.33]{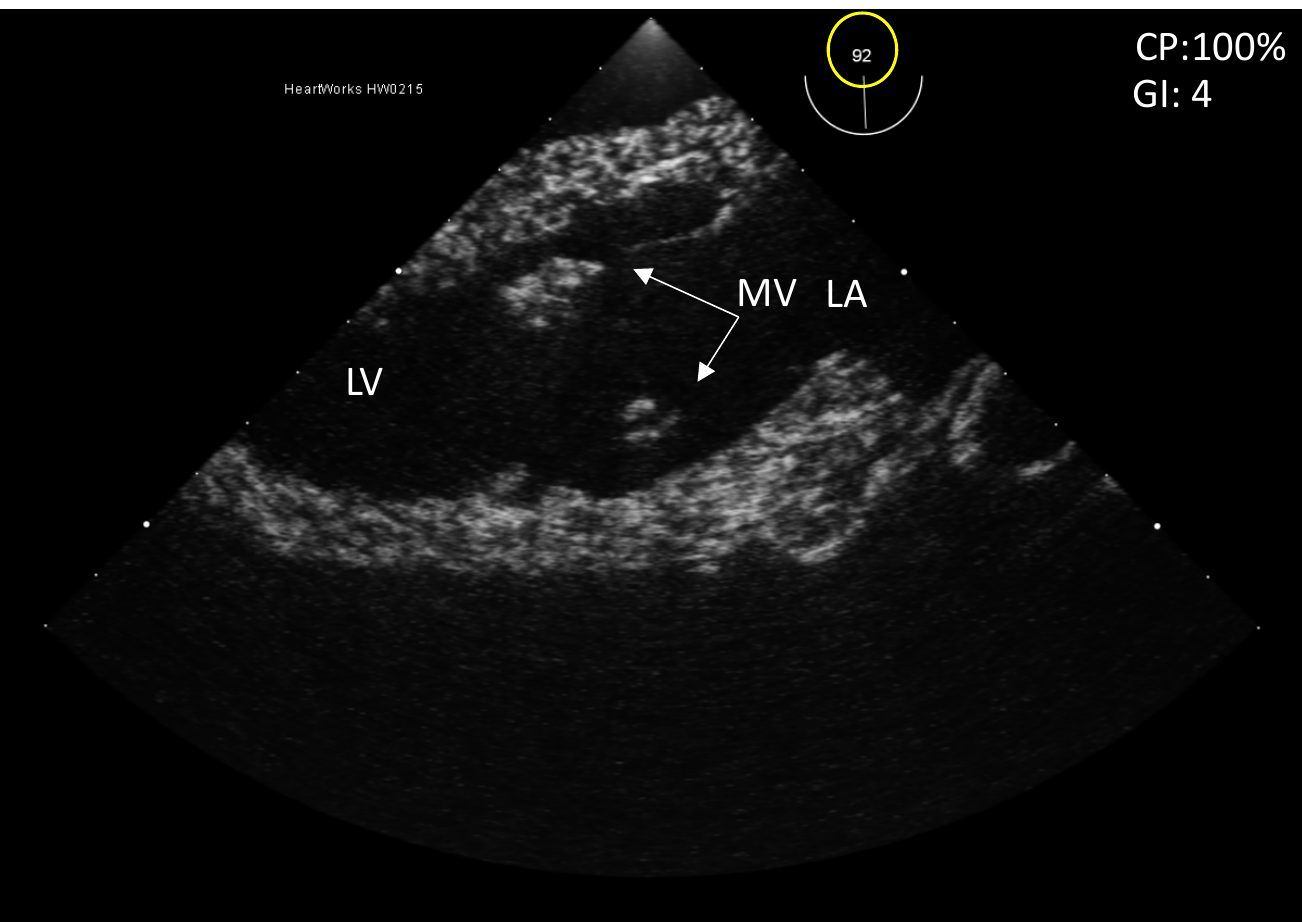}}
  \end{tabular}%
  }{%
  \captionof{figure}{Scoring examples for Views 3 and 7, from different participants, with annotated structures of importance. Left images are scored poorly whereas right images obtain excellent marks.  \textbf{Top row, View 3} - LA: left atrium, RA: right atrium, TV: tricuspid valve, RV: right ventricle, AV: aortic valve, PV: pulmonary valve, circle indicates visibility of AV cusps; \textbf{Bottom row, View 7} - LV: left ventricle, LA: left atrium, MV: mitral valve and arrows showing leaflets}\label{fig3}%
}
  \capbtabbox[0.3\textwidth]{%
    \begin{tabular}{|c|}
      \hline
      \parbox[b]{3.25cm}\centering{ \textbf{\small ME AV SAX (3)}}\\
      \hline
      \parbox[b]{3.25cm}{\footnotesize 1) 30\textdegree-45\textdegree rotation} \\
      \parbox[b]{3.25cm}{\footnotesize 2) AV centred in screen}     \\                                      
      \parbox[b]{3.25cm}{\footnotesize 3) 3 cusps visible}            \\                                     
      \parbox[b]{3.25cm}{\footnotesize 4) Imaging plane at level of leaflet tips} \\ 
      \parbox[b]{3.25cm}{\footnotesize 5) Probe tip appropriately behind LA}  \\
      \hline
    \end{tabular}
    \\[0.49cm]   
    \begin{tabular}{|c|}
      \hline    
      \parbox[b]{3.0cm}{\centering \textbf{\small TG 2C (7)}} \\ 
      \hline
      \parbox[b]{3.25cm}{\footnotesize 1) 85\textdegree-95\textdegree rotation} \\ 
      \parbox[b]{3.25cm}{\footnotesize 2) LA and LV both visible} \\ 
      \parbox[b]{3.25cm}{\footnotesize 3) MV visible on right side of screen} \\
      \parbox[b]{3.25cm}{\footnotesize 4) Post. and Ant. MV leaflets seen} \\[0.02cm]
      \hline
    \end{tabular}   
    \\[0.115cm]   
  }{%
    \captionof{table}{The checklists used for the ME AV SAX (View 3) and TG 2C (View 7) TEE views.} \label{table1}%
  }
  \end{floatrow}
  \end{figure}

In total, 38 participants of varied experience performed the experiments. The population included accredited anaesthetists having performed more than 500 exams, less experienced practitioners and trainees in the early stage of their residency. Participants were allowed time to familiarise themselves with the setup and the simulator. Manual scoring was blindly performed by three expert anaesthetists based solely on the acquired videos/images. Each view was assessed with two distinct image quality metrics. The first metric is a criteria-based score evaluated on a predetermined checklist, of which each item was assigned a binary value (0-not met, 1-met). The checklists for two of the views are depicted in \tablename\,\ref{table1} and are derived following the latest ASE/SCA imaging guidelines for each view \cite{hahn13}. This technique broadly evaluates three attributes, the correct angulation of the US probe in each view, the presence/visibility of specific heart tissue and the proper positioning of the probe in the oesophageal lumen. The number of items varied for different views so did the maximum score. The percentage of criteria (CP) met over the total number was used to provide a uniform measure among all views. The second score is a general impression (GI) assessment of the US video/image scored on a 0-4 scale, which assess the overall quality of the acquired image. Grades from the three evaluators were averaged to obtain a single mean score per US view for each volunteer. As expected the two scores are highly correlated ($\rho \sim0.93$). Inter-rater variability was independently evaluated for each view, using the interclass correlation coefficient (ICC) and Krippendorff's Alpha (KA). Both metrics show very good agreement between the three evaluators with ICC $\sim 0.9$ and KA $\sim 0.8$ for all views.

\figurename\,\ref{fig3} illustrates two examples in the opposite ends of the quality spectrum from views 3 and 7. The average quality scores are given inset and we annotated the elements in the images that satisfy the criteria in the checklist of each view, provided in \tablename\,\ref{table1}. The images on the left are of poor quality and only meet a small number of the checklists' items. For example the top left ME AV SAX image has the correct probe rotation and visualises the three cusps of the aortic valve. It fails to meet the rest of the criteria. The bottom left image of the TG2C view, only achieved correct probe angulation, but because of inadequate positioning fails to satisfy the rest of the criteria. Consequently, both CP and GI scores are low, since both images on the left side are of unacceptable quality. Images on the right side are examples of ideally imaged views fully satisfying the respective checklists and achieving full marks in both metrics.

We recorded 365 video sequences from the 38 participants with 15 views failing to store properly. For our investigation, we extracted all 16060 (i.e. 365 x 44) frames from the stored videos and used the mean manual scores as labels. All frames from a given video were labelled with the average score of that view on the premise that reviewers assigned their grades after watching the short videos so we consider that the mark equally represents all frames. No probe movement takes place in the videos, only the simulated beating of the heart model. Therefore the qualitative attributes of the stored view are the same in all frames. We divide the dataset using the 80\% - 20\% rule for training and testing, considering the total number of volunteers. Frames from 32 participants were designated for training (13464) and from 6 for testing (2596). 

\subsection{CNN Architectures}
\begin{figure}[!t]
  \begin{flushleft}
      \subfloat[]{\includegraphics[scale=0.20]{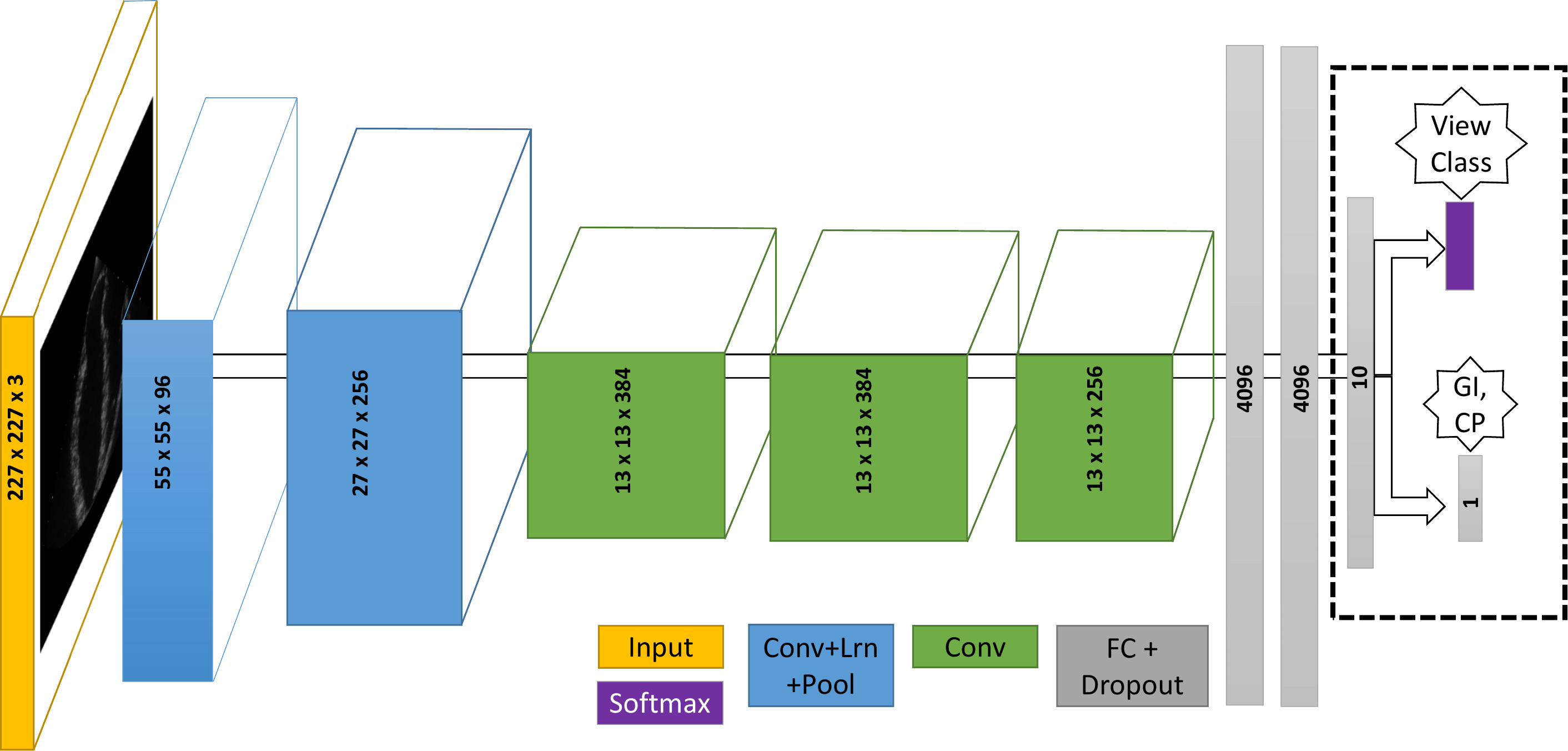}} \hfill
      \subfloat[]{\includegraphics[scale = 0.21]{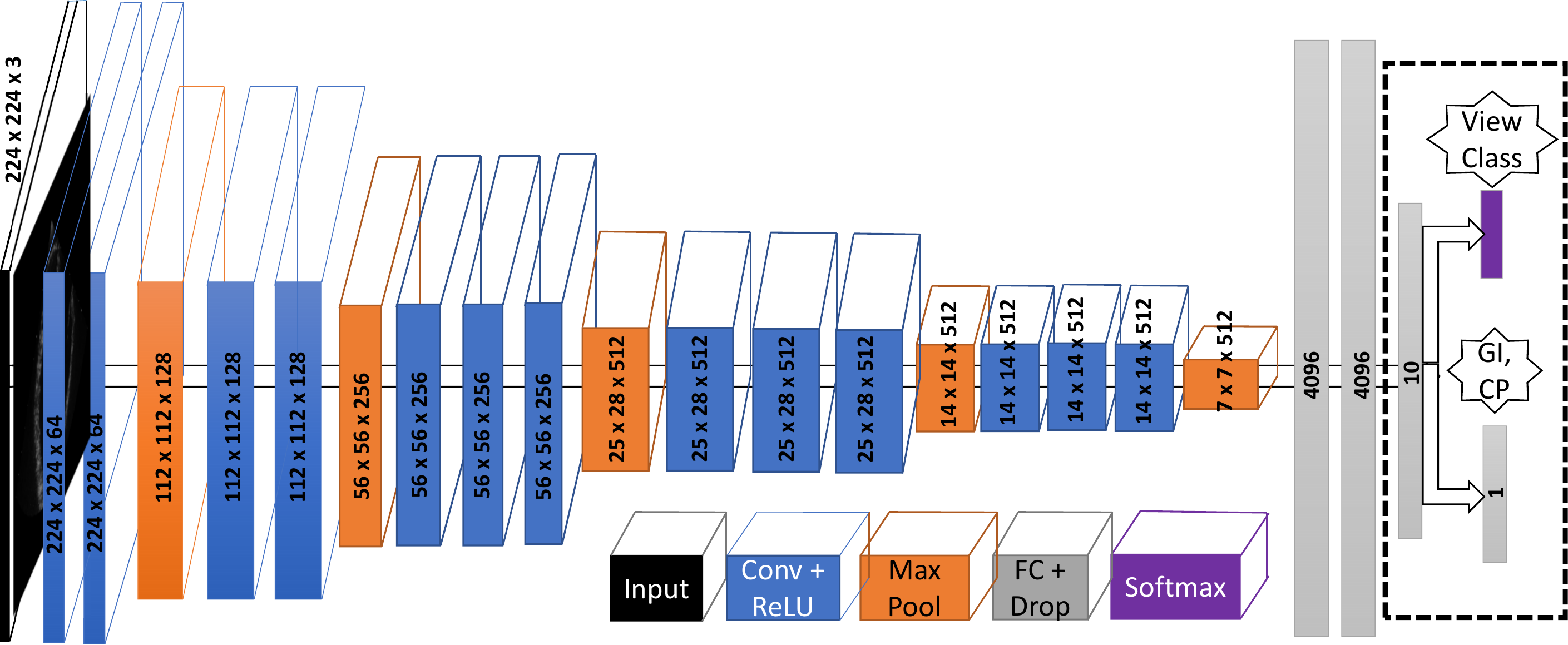}}    
      \caption{The two networks (a)Alexnet, (b)VGG, developed for the TEE score estimation task. The customized output stage with the added FC layers and the softmax activation for classification is enclosed in the boxes.}
      \label{fig4}  
  \end{flushleft}
\end{figure}

We opted to develop CNN models for performing a regression task and train them to learn to estimate the performance score as a single continuous variable; $CP\in \{0, \dotsc, 100\}$, $GI\in \{0, \dotsc 4\}$. Since the checklists' criteria and their number are different among views, it was not feasible to structure and train a single model for evaluating individual criteria for all views. This would require a non-efficient approach with separate sub-models per view. Hence a single CP score per view was computed and estimated. We experimented with two established CNN architectures namely, Alexnet and VGG, originally built to perform image classification tasks \cite{krizhevsky12, simonyan14}. We repurpose them by restructuring their output stage and consider 10 available classes, one for each TEE view. The final fully-connected (FC) layer of both CNNs is resized with a dimension of 10. One additional FC layer with output size $d=1$ and linear activation is added to complete the regression operation and estimate the score. For classifying the input to one of the TEE views, softmax activation is applied after the FC layer with $d=10$. Effectively we structure our network so that it can be trained to both estimate the performance scores and recognize the corresponding view of the input. \figurename\,\ref{fig4} illustrates the two customised architectures with the added layers.

\section{Experimentation and results}
\label{sec3}
\begin{figure*}[!ht]
  \begin{tabular}{cccc}
  \subfloat{\includegraphics[scale=0.20]{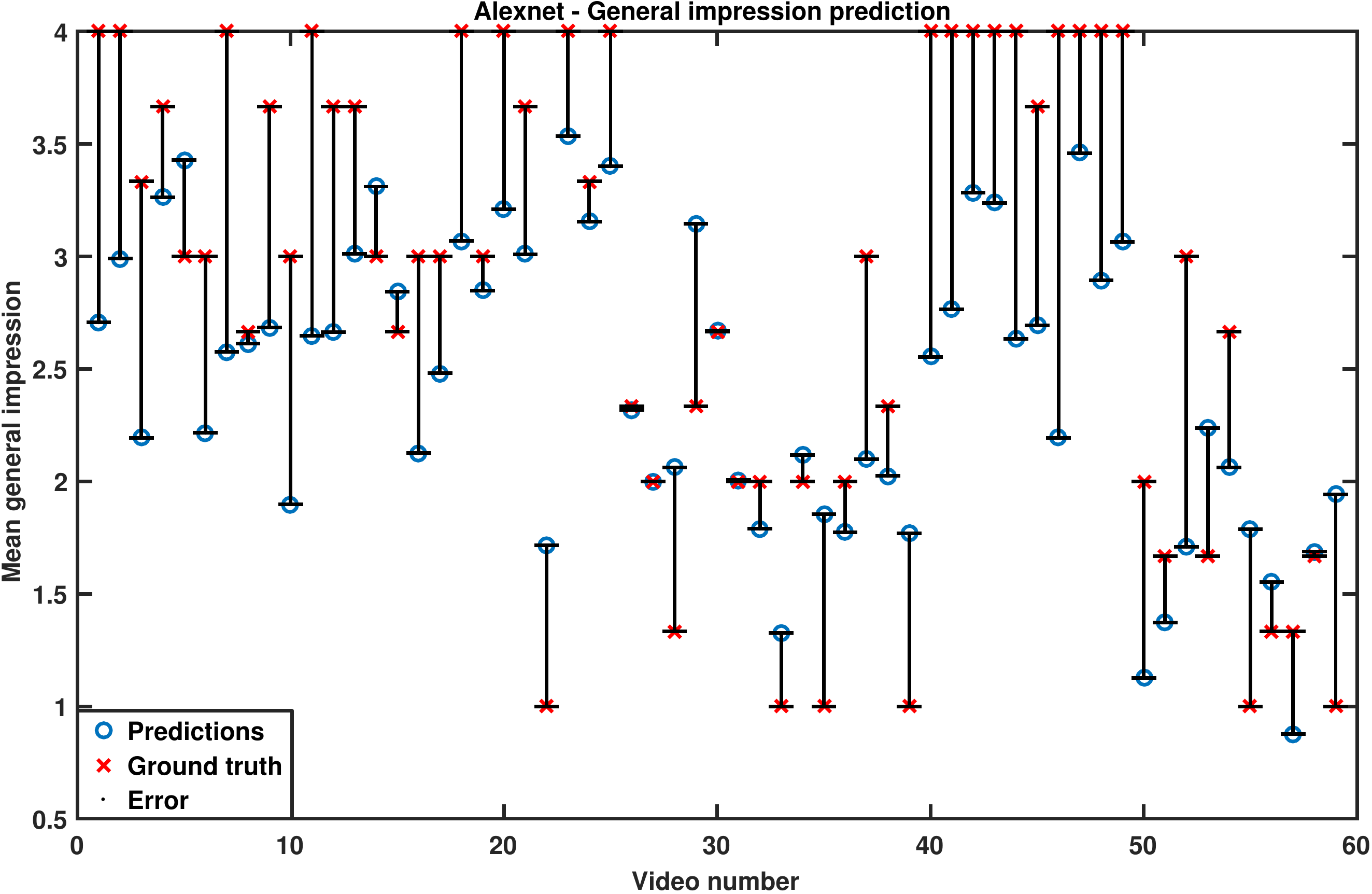}}  & \subfloat{\includegraphics[scale=0.20]{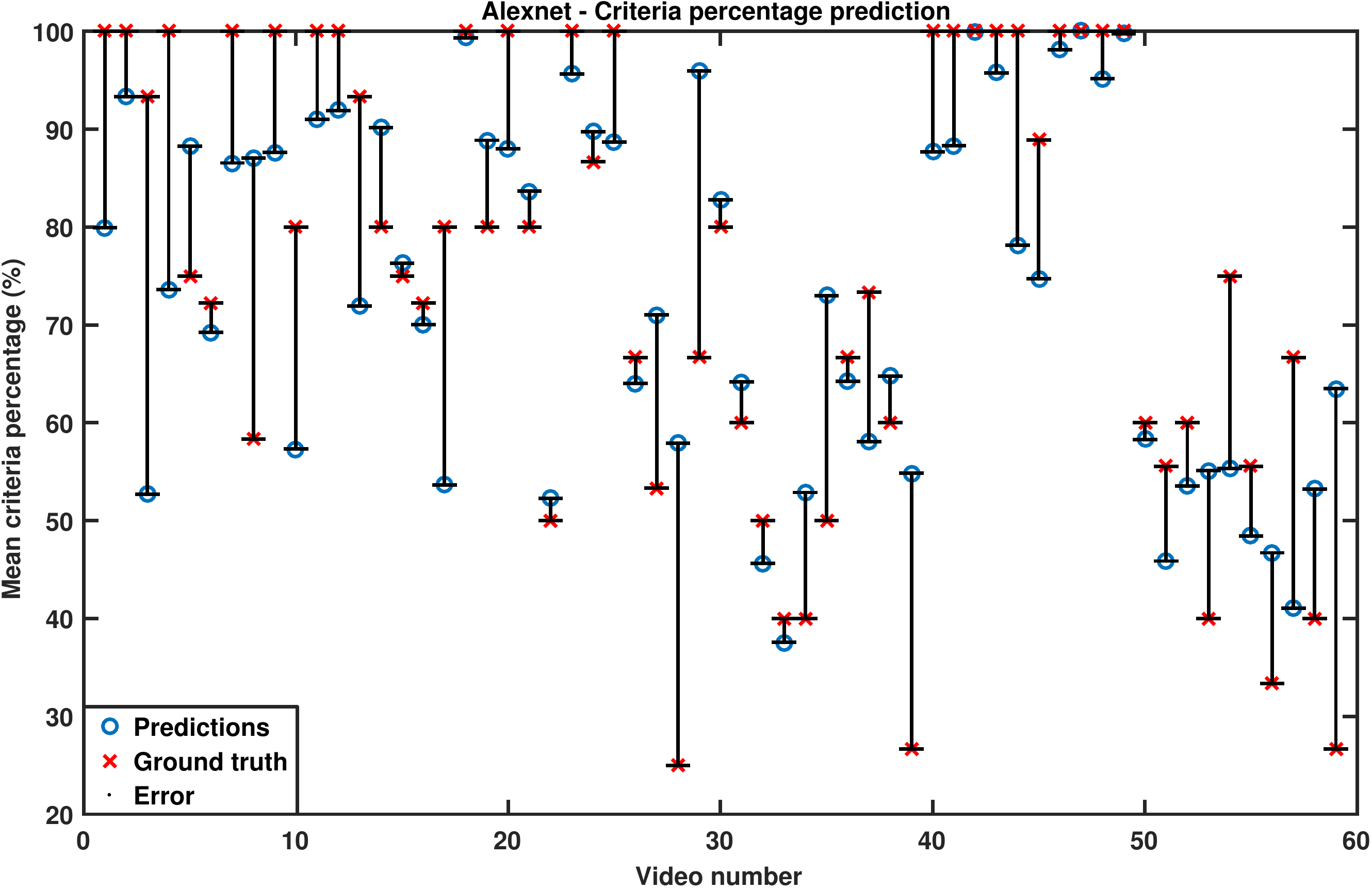}}\\ 
  \subfloat{\includegraphics[scale=0.20]{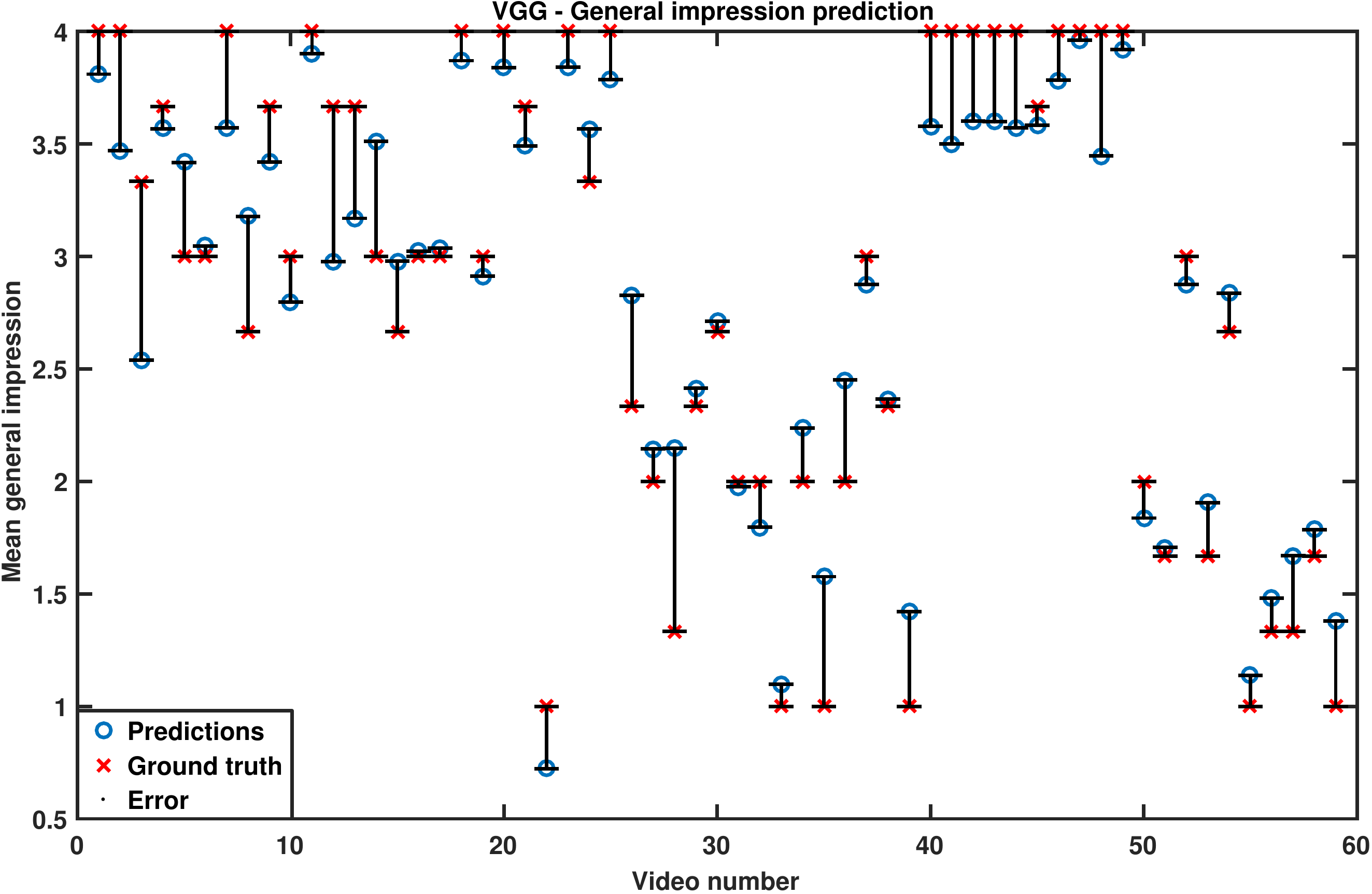}}  &  \subfloat{\includegraphics[scale=0.20]{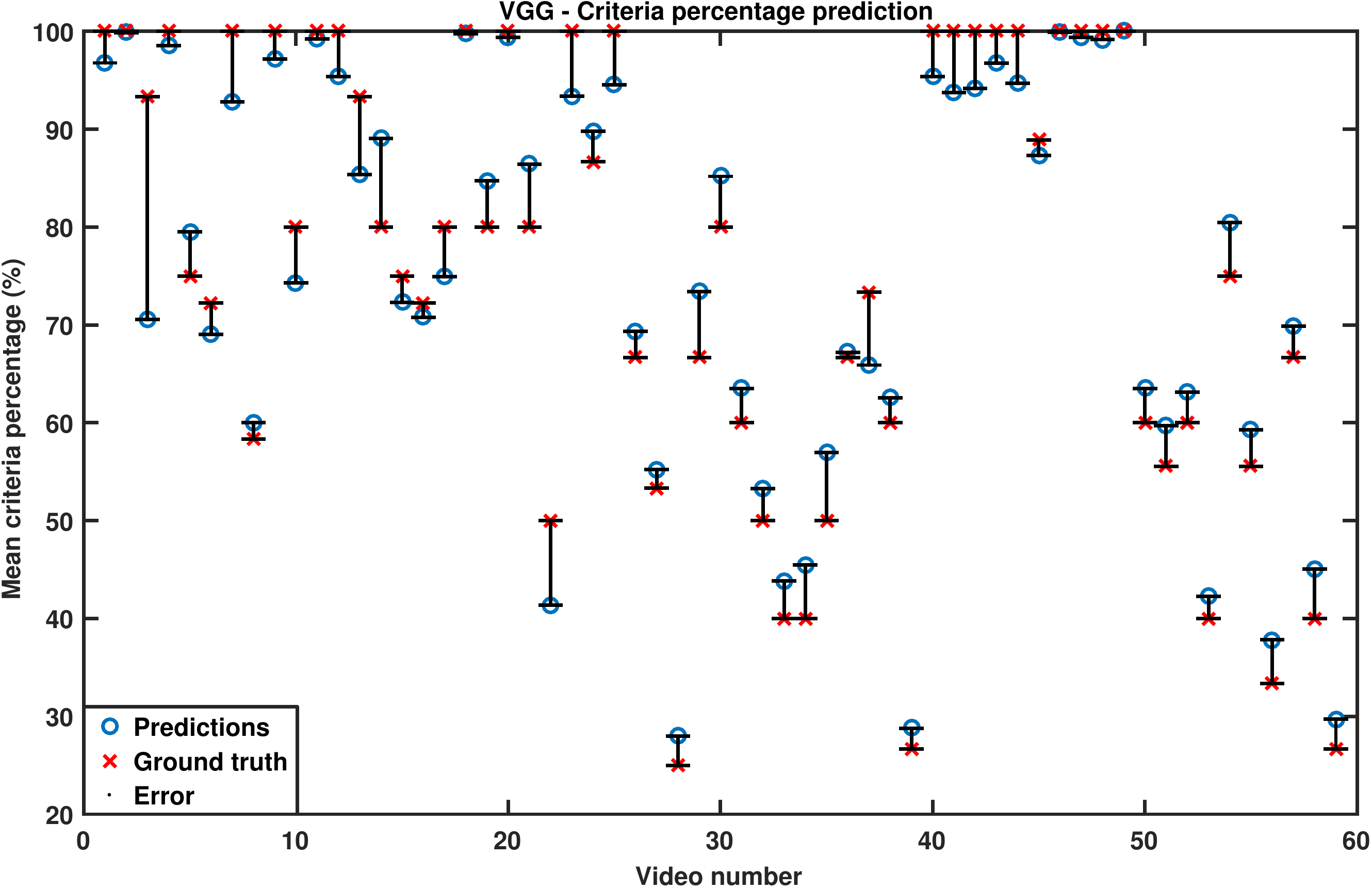}}
  \end{tabular}
  \caption{Grouped estimation results per testing video. RMSE and average $\sigma$ values: (top) Alexnet - CP: 15.78 ($\sigma$=3.34), GI: 0.8 ($\sigma$=0.22); (bottom) VGG - CP: 5.2 ($\sigma$=3.55), GI: 0.33 ($\sigma$=0.23)}
 \label{fig5}
\end{figure*}

CNN models were implemented with the TensorFlow framework. The training dataset was randomized and images were resized from 1200x1000, to 227x227 for Alexnet and 224x224 for VGG. Batches of 128 (Alexnet) and 64 (VGG) were used. The mean square error was set as the loss function and gradient descent optimization with adaptive moment estimation was performed with a learning rate of 0.001. Both networks were initialized with publicly available weights from the ILSVRC challenge \cite{krizhevsky12, simonyan14}, apart from the additional dense layers we introduced, which were assigned random weights and trained from scratch. Backpropagation was used to update the weights. The two architectures were independently trained for each performance metric and convergence was achieved after 2K iterations for Alexnet and after 12K for VGG. The models were also trained to classify images to their respective view, achieving over 98\% accuracy. 
\begin{table}[!t]
  \caption{Overall and interval RMSE results of the developed networks.}
    \begin{tabular}{|@{}c@{}|c|c|c|c|c|@{}c@{}|}
    \hline
    \multicolumn{6}{|c|}{Criteria percentage score (CP)} \\
    \cline{1-6}
    \parbox{1.6cm}{\centering \textbf{Network}}  &  $CP < 55\%$  & $55\% \geq CP < 75\%$  & $75\% \geq CP < 90\%$  & $CP \geq 90\%$  & \textbf{Total}\\
    \hline
   Alexnet  & 20.38 & 14.59 & 18.9 & 12.1  & 16.23\\
    \hline	
    VGG & 5.55 & 5 & 11.8 & 5.34 & 7.28\\
    \hline
    \multicolumn{6}{|c|}{General impression score (GI)} \\
    \cline{1-6}
    \parbox{1.5cm}{\centering \textbf{Network}}  &  $GI < 1.8$  & $1.8 \geq GI <2.8$ & $ 2.8\geq GI <3.8 $  & $ GI > 3.8 $ & \textbf{Total}\\
    \hline
   Alexnet  & 0.65 & 0.44 & 0.84  & 1.13 &  0.83 \\
    \hline	
    VGG & 0.42 & 0.31 & 0.46 & 0.45 & 0.42\\
    \hline
    \end{tabular}
  \label{table2} 
\end{table}
\tablename\,\ref{table2} lists overall RMSE results and the RMSE on score intervals, from estimating the two image quality scores on the 2596 testing images. Both models perform adequately but,owing to its denser structure, VGG outperforms Alexnet significantly and has smaller error variability, providing excellent accuracy for both metrics. To obtain a single score per video, similarly to the three evaluators, we grouped the predictions of the frames from the same video and averaged them. The per video results, for 59 videos from the 6 testing participants (one video was not captured) are shown in \figurename\,\ref{fig5}. The RMSE of the grouped results is lower for both networks, that also give consistent estimations in frames from the same video, indicated by low standard deviation values ($\sigma_{CP}\simeq3.5$, $\sigma_{GI}\simeq0.2$). 

\section{Conclusions}
\label{sec4}
In this article we demonstrated the applicability of CNNs architectures for automated quality evaluation of TEE images. We collected a rich dataset of 16060 simulated images graded with two manual scores (CP, GI) assigned by three evaluators. We experimented with two established CNN models, restructured to perform regression and trained these to estimate the manual scores. Validated on 2596 images, the developed models estimate the manual scores with high accuracy. Alexnet achieved an overall RMSE of 16.23\% and 0.83, while the denser VGG had better performance achieving 7.28\% and 0.42 for CP and GI respectively. These very promising outcomes indicate the potential of CNN methods for automated skill assessment in image-guided surgical and diagnostic procedures. Future work will focus on augmenting the CNN models and investigating their translational ability in evaluating the quality of real TEE images. 

\subsection*{Acknowledgements}
The authors would like to thank all participants who volunteered for this study. The work was supported by funding from the EPSRC (EP/N013220/1, EP/N027\\078/1, NS/A000027/1) and Wellcome (NS/A000050/1). 
\bibliographystyle{splncs04}

\end{document}